\documentclass[]{article} 
\usepackage{arxiv} 
\usepackage{apacite}
\usepackage{times}  
\usepackage{helvet}  
\usepackage{courier}  
\usepackage{url}  
\usepackage{graphicx}  
\usepackage{tabularx}
\usepackage{makecell}

\begin{document}
\title{On the Semantic Interpretability of \\ Artificial Intelligence Models}
\author{Vivian S. Silva\textsuperscript{1}, Andr\'{e} Freitas\textsuperscript{2}, Siegfried Handschuh\textsuperscript{1,3} \\
	\textsuperscript{1}Department of Computer Science and Mathematics, University of Passau,
	Germany \\
	\textsuperscript{2}School of Computer Science, University of Manchester, UK \\
	\textsuperscript{3}Institute of Computer Science, University of St. Gallen, Switzerland \\
	vivian.santossilva@uni-passau.de, andre.freitas@manchester.ac.uk, siegfried.handschuh@unisg.ch}

\date{}

\maketitle

\begin{abstract}
Artificial Intelligence models are becoming increasingly more powerful and accurate, supporting or even replacing humans' decision making. But with increased power and accuracy also comes higher complexity, making it hard for users to understand how the model works and what the reasons behind its predictions are. Humans must explain and justify their decisions, and so do the AI models supporting them in this process, making semantic interpretability an emerging field of study. In this work, we look at interpretability from a broader point of view, going beyond the machine learning scope and covering different AI fields such as distributional semantics and fuzzy logic, among others. We examine and classify the models according to their nature and also based on how they introduce interpretability features, analyzing how each approach affects the final users and pointing to gaps that still need to be addressed to provide more human-centered interpretability solutions.
\end{abstract}

\section{Introduction}
Artificial Intelligence is becoming a ubiquitous presence in our everyday lives. Supporting and expanding the cognitive abilities of humans or even replacing them, powerful algorithms along with huge amounts of data can now perform a wide variety of tasks, from labeling images to predicting cancer, as well as or even better than a human would do. As AI models grew in processing power and accuracy, they also became more complex, and their predictions, as accurate as they may be, don't bring with them a clear explanation on how they were achieved. Such operation paradigm may bring drawbacks with it because, as it was recently reinforced \cite{kuang2017can}, AI must ``conform to the society we've built -- one in which decisions require explanations, whether in a court of law, in the way a business is run or in the advice our doctors give us''. This leads us to the need for moving from simply accepting \textit{what a model does}, to interpreting it to understand \textit{how it does} so.

Interpreting the model behavior is not always necessary. Users probably won't want further explanation about the outputs of systems performing voice recognition or image classification, as long as they behave reasonably as expected. But, even in those cases, understanding how the model works can sometimes be of great help. Back in 2015, Google was embarrassed by its Photos service labeling pictures of black people as ``gorillas'' \cite{simonite2018when}. Why was the image recognition software doing so? Certainly not because it is racist and deliberately meant to offend people, but rather because it was making the wrong correlations between the features extracted from the pictures. Google's answer to the problem was to ban the words ``gorilla'', ``chimp'', ``chimpanzee'', and ``monkey'' from the Photos lexicon altogether, avoiding people (and monkeys) being assigned such labels, and this workaround remains in place, almost three years on. Although there may be many different reasons for Google's approach towards the issue, it becomes clear that the complexity of the machine learning algorithm employed for image recognition may have prevented a proper quick fix, that is, it wasn't possible to quickly interpret the model, identify, and adjust its malfunctioning parts in a timely manner.

However, misclassifications can cause much more than annoyance. The greatest importance of model interpretability rests on decision-making systems, whose outputs can have a material impact on the lives of individuals. Artificial intelligence techniques are now being largely used in tasks such as medical diagnosis, insurance and credit assessment, and criminal recidivism prediction, among others. In those cases, even though a system is known to make accurate predictions, explaining and justifying these predictions may be crucial for users to trust it and make further decisions based on these outputs. Same Google has just released a new AI algorithm capable of predicting heart diseases by analyzing data generated from scans of the back of patients' eyes \cite{poplin2018prediction}. This algorithm can make the assessment of a patient's cardiovascular risk quicker and easier, but, although all authors say is that it still needs to be thoroughly further tested before being used by doctors, it also needs to be interpretable: it must make clear what information from the scans and what correlations between them are leading to a diagnosis, so doctors can have the necessary evidence to judge whether to follow the system's recommendation or not. Moreover, relying on such technology for prescribing medical treatments won't allow for a quick workaround in case the model start showing undesirable behavior, as in the image classification scenario.

The importance of the so-called Explainable AI lies not only on the need for evidence to support decision making but also on the demand to easily identify biased correlations that could go unnoticed otherwise. Zhao et al. \citeyear{zhao2017men} argues that many prediction models risk reflecting social biases found in data, showing that, using an image dataset containing significant gender bias where the activity ``cooking'' was over 33\% more likely to involve women than men, a trained model further amplifies the disparity to 68\% at test time. Transported to the decision-making scenario, sensitive information such as gender, race, religion or income, for example, can lead to unfair predictions on tasks that involve individual profiling, such as hiring, loan granting or crime prediction, to name a few, where certain groups can be subject to discrimination. Even though the model's predictions may seem to conform to previous decisions, if those decisions were influenced by social biases and this is reflected in the data they generated, an interpretable model can make that clear and allow for more fairness on future verdicts.

Although semantic interpretability is gaining renewed attention due to the increasing use of machine learning, being interpretable is not an issue exclusive to these models, but a requirement for any approach dealing with AI. For example, in an evaluation challenge asking participants to rate the semantic similarity of pieces of texts and explain their decisions \cite{agirre2015semeval}, approaches as different as knowledge base search \cite{hanig2015exb,hassan2015fcicu,banjade2015nerosim}, rule-based \cite{banjade2015nerosim,karumuri2015umduluth}, referential translation machine \cite{biccici2015rtm}, and support vector machine \cite{agirre2015ubc} were employed to implement an interpretable text analysis system. Interpretability is an AI concern rather than purely a machine learning matter. Therefore, in this work we seek to offer a broader view on interpretability, analyzing the efforts of different types of AI models to become more interpretable and how the concept of interpretability is dealt with by each of them.

We start by examining the concept of interpretability itself: how it is regarded across different fields and what shapes it can assume. We then analyze several models that claim to be interpretable and the evaluation methods and initiatives intended to measure a model's level of interpretability. Finally, we look at the human-centric aspect of semantic interpretability, classifying models according to how they implement and what they offer as interpretations and pointing to gaps that still need to be addressed to meet explanation requirements from the final user point of view. The list of analyzed models is by no means exhaustive; we sought to pick representative examples of each class, focusing on the ones that emphasize and prioritize interpretability as a driving design choice in the model construction.

\section{Interpretability across Models}
Interpretability issues have been gaining the spotlight in the recent years due to the fast advancements and widespread utilization of machine learning techniques, especially the ones based on deep neural network models. Such models proved to be powerful predictors, but its complexity usually prevents the user from understanding its internal dynamics. To trust supervised machine learning models we need them to be not only accurate, but interpretable \cite{lipton2016mythos}. However, the need for interpretability is not exclusive to machine learning models. Rule-based models can also grow in size and complexity to a point where users are similarly left unable to comprehend them, also requiring interpretability issues to be taken into account, so that keeping track of those models' decisions becomes feasible. In fact, interpretability has been a key issue in many different areas of AI for many years, notably in the design of fuzzy logic models, giving origin to a number of theoretical and practical studies regarding this topic \cite{alonso2011special,mencar2011interpretability,alonso2014some}.

Being interpretable is sometimes regarded as an inherent attribute of the model. Kotsiantis \citeyear{kotsiantis2007supervised}, for example, states that logic-based algorithms such as Na\"{\i}ve Bayes, decision trees, and rule learners are naturally easy to interpret, while neural networks, SVMs and K-NNs have very poor interpretability. Nevertheless, this is not always accepted as a fact. Lipton \citeyear{lipton2016mythos} questions this assumption arguing that ``neither linear models, rule-based systems, nor decision trees are intrinsically interpretable'', adding that ``sufficiently high-dimensional models, unwieldy rule lists, and deep decision trees could all be considered less transparent than comparatively compact neural networks''. This suggests that interpretability is a property that must be pursued rather than being taken for granted as a result of the model choice.

But what could, in fact, be called \textit{interpretability}? Being tackled in the scope of different approaches, it is natural that the definition of ``interpretable model'' is not yet something uniformly agreed upon. Lipton \citeyear{lipton2016mythos} observes that ``both the motives for interpretability and the technical descriptions of interpretable models are diverse and occasionally discordant, suggesting that interpretability refers to more than one concept''. He argues that the purpose of an interpretation is to convey useful information and that this can be done even without shedding light on a model's inner workings. This leads to the division of interpretability techniques into two broad categories: one referring to \textit{transparency}, asking \textit{how the model works}, and the other relating to \textit{post-hoc explanations}, inquiring \textit{what else the model can tell} \cite{lipton2016mythos}. That means that interpretability can be related to the system output, but also to the system architecture itself.

Most of the models that claim to be interpretable focus on \textit{transparency}, that is, given the input data and the model parameters, it should be possible to step through the calculations that lead to a prediction. Transparency refers not only to the model as a whole (\textit{simulatability}), but also to each of its parts: each input, parameter, and calculation should admit an intuitive explanation (\textit{decomposability}) \cite{lipton2016mythos}.

\textit{Post-hoc explanations}, on the other hand, can be seen as an abstraction layer for the model, as they can summarize and translate the system's procedures into a friendlier format, exempting the user from going through algorithmic details. Natural language explanations, visualizations of learned representations or models, and explanations by example (such as presenting the k-nearest neighbors of a word given its vector representation) are some common semiotic approaches for providing post-hoc model interpretation \cite{lipton2016mythos}. Explanations are especially necessary when the problem formalization is incomplete, due to, for example, the infeasibility of predicting all possible outputs given all possible inputs, or the abstract nature of some system requirements, such as \textit{fairness} and \textit{trust} \cite{doshi2017roadmap}. In an incompleteness scenario, explanations are a resource to make possibly flawed results (and their causes) clearly visible, allowing users to act on them.

Biran and Cotton \citeyear{biran2017explanation}, focusing on machine learning, also distinguishes between two research branches, which they call \textit{interpretable models}, equivalent to the transparent models in Lipton's classification; and \textit{prediction interpretation and justification}, that is, the previously seen post-hoc explanations. However, they limit the usage of the term \textit{interpretability} only to the understandability of the model's internal operations, arguing that a system can provide justifications (as the output) without being interpretable, that is, without making clear to the user what its internal procedures are.

The importance of both varieties of interpretability can be noted in the European Union's new General Data Protection Regulation, to take effect as law across the EU in 2018. It covers two points tightly related to interpretability: the \textit{non-discrimination} in automated individual decision-making and the \textit{right to explanation} \cite{goodman2016eu}. The first point refers to algorithmic transparency: systems that support decision-making based on individual profiling, such as credit and insurance assessment platforms, should ensure that they do not produce discriminatory results by using variables coding for race or ethnicity, income, or any other sensitive information. That means systems must make clear not only what information they are using but also what correlations algorithms are extracting from data for making predictions. The second point is directly related to the systems' ability to provide (post-hoc) explanations and justify decisions reached after algorithmic assessment in a human-understandable way.

Regardless of the chosen technique to render a model interpretable, the most important aspect to be kept in mind is that interpretability is a human-centered feature. Doshi-Velez and Kim \citeyear{doshi2017roadmap} define interpretability as ``the ability to explain or to present in understandable terms to a human'', while Alonso et al. \citeyear{alonso2015interpretability} observe that ``the importance of the human component implicitly suggests a novel aspect to be taken into account in the quest for interpretability'', both emphasizing that the aspects of human cognition should be put at the center of modeling decisions. Offering an interpretation of a model can be seen as a knowledge extraction process, and as such it must take into account the human cognitive factor it inherently involves \cite{vellido2012making}.

Some studies have tried to draw the users' preferences when it comes to interpretation, especially explanations. Miller et al. \citeyear{miller2017explainable} summarize the findings pointing that people usually judge explanations based on \textit{pragmatic influences} of causes, which include usefulness and relevance, among others, rather than the probability that the cited cause is actually true. Also, people prefer explanations that are \textit{simpler} (cite few causes), more \textit{general} (they explain more events), and \textit{coherent} (consistent with prior knowledge), favoring simpler explanations over more likely explanations. They conclude that ``giving simpler explanations that increase the likelihood that the observer both \textit{understands} and \textit{accepts} the explanation may be more useful to establish trust, if this is the primary goal of the explanation'' \cite{miller2017explainable}.

As important as defining what interpretability \textit{is} is understanding what \textit{it is for}. Models, and consequently the systems they support, are ultimately designed to address some human user need, and making them interpretable is intended to ensure that the users' requirements are met in the sense that they can use, understand and trust the system in the simplest and most effective way possible.

\section{Interpretability-driven AI Models}
In this Section, we review different types of models which emphasize interpretability.  We focus on models that were explicitly designed to be interpretable, that is, models where introducing or increasing the interpretability was a primary requirement driving the design choices. We divide the models into three categories: \textit{data models}, \textit{algorithmic models}, and \textit{hybrid models}, which take into account the content where an interpretation is to be extracted from. These categories are not mutually exclusive: algorithmic models can have minor internal data representations and vice-versa; therefore for classifying the models we considered the most predominant characteristic of each of them.

\subsection{Data Models}
Interpretable data models are models whose core component is preprocessed data, which goes through some clusterization process for posterior use. Although these models are usually built by some machine learning method, what define such models, rather than the construction procedure, is the shape and content of the final product, which will be used as input for a number of other tasks.

The most outstanding examples of such models are interpretable \textbf{Distributional Semantics Models (DSMs)}. These models are grounded in the distributional hypothesis, which states that words that occur in similar contexts tend to have similar meanings \cite{turney2010frequency}, and allow words to be represented as a vector summarizing their patterns of co-occurrence in large text corpora. The vector representations usually go through a dimensionality reduction process, a mathematical operation that makes the vectors more manageable while still capturing the co-occurrence patterns \cite{baroni2010strudel}, being the most common technique the Singular Value Decompositions (SVD) \cite{klema1980singular}.

Dimensionality reduction results in vectors whose features correspond to very broad domains of knowledge, such as ``food'', ``sports'' or ``education'', for example. A direct consequence of this new representation, as observed by Baroni et al. \citeyear{baroni2010strudel}, is that the underlying abstraction behind most DSMs, the Vector Space Models (VSMs), ``might be very good at finding out that two concepts are similar, but they tell us little about the internal structure of concepts and, hence, why or how they are similar'' \cite{baroni2010strudel}. What can be obtained is an overall similarity score that does not convey any additional information about the relationship between similar words. SVD, in particular, produces matrices where, for most dimensions, it is hard to interpret what a high or low score entails for the semantics of a given word \cite{fyshe2015compositional}. This is illustrated in the example posed by Murphy et al. \citeyear{murphy2012learning}, where they retrieve the latent dimension of an SVD-based model for which the word ``pear'' has its largest weighting and whose most strongly positively associated tokens are ``action'', ``records'', ``government'', ``record'', and ``search''. As can be seen, it is not clear at first sight what the relationships between the words in this dimension are or even to what semantic category they all belong in, making it hard to extract an interpretation from it.

To overcome this problem, some approaches for building more interpretable DSMs have been proposed. An example is Strudel (structured dimension extraction and labeling), a corpus-based semantic model that induces semantic information from naturally occurring data using part-of-speech (POS) tagging, lemmatization of the corpus, and a set of extraction templates defined over POS sequences. The model's main goal is to extract dimensions which are ``interpretable as properties, automatically annotated with information about the nature of the relation they instantiate'' \cite{baroni2010strudel}. That means it involves some relation extraction functionality, but instead of being predefined, the relations are inferred from the co-occurrence patterns, that is, from the distribution of patterns connecting a concept to its properties. For example, for the concept ``book'', Strudel associates the following properties, along with the correspondent relations (expressed by either verbs or prepositions): they ``are written'', ``published'', and ``read'', they are ``by an author'', ``from a publisher'', ``for a reader'', and ``on a subject'', they ``have pages'' and ``chapters'', and they ``are in libraries''. When compared with speaker-generated descriptions, the property-based concept representations produced by Strudel showed to be reasonable both quantitatively and qualitatively.

Murphy et al. \citeyear{murphy2012learning} present a technique called Non-Negative Sparse Embedding (NNSE) for learning interpretable distributional semantic models. They define interpretability from the point of view of cognitive plausibility, stating that a word representation is interpretable if each of its dimensions is semantically coherent. They measure this coherence through the \textit{word intrusion detection} task, in which, for each dimension, a set is created containing its top five words and an \textit{intruder} word. The sets are then presented to human evaluators who need to identify the intruder. A high precision in this task means the dimension is interpretable because the human evaluator can easily name the category it is representing and pick out the word that is not a member of this category, i.e., the intruder. One example of such sets is the one composed of the words \{``bathroom'', ``closet'', ``attic'', ``balcony'', ``quickly'', ``toilet''\}; here it is easy to name the category as ``house parts'' and point ``quickly'' as the intruder.

CNNSE (Compositional NNSE) is a variation of NNSE intended to allow word and phrase vector to adapt to the notion of composition by learning a DSM that supports semantic composition operations \cite{fyshe2015compositional}. CNNSE phrasal vector representations have shown to be a better match to actual phrase meanings when judged by human evaluators. JNNSE (Joint NNSE) is another NNSE variation that combines the representations obtained from large text corpora with brain activation data recorded while people read words \cite{fyshe2014interpretable}.

The Explainable Principal Component Analysis (EPCA) \cite{brinton17framework} is yet another technique for generating interpretable vector representations. It builds upon the Principal Component Analysis (PCA) dimensionality reduction approach \cite{jolliffe1986principal}, including a human-in-the-loop stage for refining the data. The EPCA process is performed iteratively: basis vectors generated by the regular PCA are analyzed by a (human) model designer, who excludes any word not related to the general category implied by all the other words in the vector, creating the first explainable basis vector. This vector is then excluded from the input data, over which regular PCA followed by the manual procedure are performed again, generating the second explainable basis vector. The process goes on until all the possible explainable basis vectors have been identified. Although the human curation can doubtless improve the model interpretability, the author presents no study regarding the approach feasibility. Relying on such amount of human interaction could be a prohibitive costly task, given that vector representations are usually built over very large corpora.

DSMs are widely used as input features for machine learning models addressing a large variety of tasks. Using more coherent and interpretable data models can potentially increase the interpretability of the algorithmic models using them as input, providing an additional source of information for the generation of post-hoc explanations. We discuss interpretable algorithmic models in the next Section.

\subsection{Algorithmic Models}
Interpretable algorithmic models are models which work by executing a sequence of computations over data, possibly using a set of parameters, to perform a given task. Different from data models, what matters here is not what they \textit{are composed of}, but rather \textit{how they work}.

\textbf{Fuzzy logic} is an example of a field of study where interpretability has long been a central concept. As observed by Alonso et al. \citeyear{alonso2011special}, ``thanks to their semantic expressivity, close to natural language, fuzzy variables and rules can be used to formalize linguistic propositions which are likely to be easily understood by human beings''. But they also point out that, besides being a feature taken for granted even inside the fuzzy community, interpretability is not an intrinsic property of fuzzy models. Although fuzzy logic has a natural inclination to interpretability, whether every element in a fuzzy system can be checked and understood by a human being heavily depends on how the system is designed \cite{alonso2011special}.

Mencar et al. \citeyear{mencar2011interpretability} define fuzzy model interpretability as a relation between \textit{fuzzy sets} -- the basic elements of a fuzzy rule base -- and \textit{concepts} -- basic units of human knowledge. Fuzzy sets and concepts are linked by the common linguistic terms they refer to, so a fuzzy model can be said interpretable when its explicit semantics, that is, the linguistic representation of fuzzy sets, is cointensive with its implicit semantics inferred by a human, i.e., the meanings they infer while reading the linguistic representation of the rules \cite{mencar2011interpretability}.

As the knowledge extracted from data by fuzzy systems must be usually communicated to users, the fuzzy community has been, in recent years, taking into account interpretability issues as a major research concern \cite{alonso2014some}. Bal\'{a}zs et al. \citeyear{balazs2013stochastic} proposed an approach based on meaning preservation (MP) -- having a common vocabulary with the user, by using linguistic terms in the same sense as the user employs them -- and a parameterizable search space narrowing method aimed at adjusting the trade-off between interpretability and accuracy commonly observed in fuzzy systems. Interpretability is not measured, but rather regarded as a binary feature: if the resulting rule base meets the predefined interpretability conditions then it is referred to as a valid interpretable solution.

Mencar et al. \citeyear{mencar2013design} argue that the fulfillment of many interpretability constraints (distinguishability, coverage, special elements, etc. \cite{mencar2008interpretability}) is guaranteed if Strong Fuzzy Partitions (SFPs) are adopted. A \textit{fuzzy partition} of the data feature is the collection of the fuzzy sets associated with each linguistic term in the model. The authors propose an approach for defining interpretable SFPs based on \textit{cuts}, points of separation between clusters within the data. Again, interpretability is not measured but considered as a natural outcome of the fuzzy partition-based adopted modeling technique.

Visual representation of rules is another resource employed to allow the interpretation of a fuzzy model. Pancho et al. \citeyear{pancho2013interpretability} describe a technique to present fuzzy association rules (FARs) to users in a graph format. Association rules identify and represent dependencies between data items in a dataset. Those dependencies are graphically depicted through fuzzy inference-grams, or \textit{fingrams}, which are networks where nodes represent fuzzy rules and the weighted edges represent interactions between rules. Nodes are always labeled with relevant textual information and the edge weighting naturally leads to the formation of distinguishable groups of rules, each associated with some value for a given variable. Although this graphical representation clearly facilitates the visual analysis and comprehension of fuzzy association rules \cite{pancho2013interpretability}, no evaluation was carried out, and the assessment of its usefulness through user feedback remains to be done.

Going beyond the theoretical aspects explored by most of the research in the area, Riid et al. \citeyear{riid2013determination} proposed a practical application to take advantage of an interpretable fuzzy model. They employ a fuzzy classifier to explain the geographical variation of Estonian folk songs metrical features. Hierarchical Clustering (HC), an agglomerative procedure based on the idea that objects tend to be more related to nearby objects than to objects farther away, is used to determine the cluster of geographical regions that show similarities regarding the verse metre. A data-driven fuzzy classifier is then used for analyzing the clusters in order to identify which features of the verse metre are critical in cluster assignment \cite{riid2013determination}. The result is a clear separation of the geographical regions into three well delineated groups but, since no evaluation was performed, it is unclear to what extent the comprehension of how the features lead to the cluster separation is, in fact, interpretable from the user point of view.

Another practical application was presented by Conde-Clemente et al. \citeyear{conde2013interpretable}. A prototype that allows a person with visual disabilities to take their own profile photos was implemented as a fuzzy control system including human-in-the-loop using natural language. The authors argue that, by making use of an approach known as Highly Interpretable Linguistic Knowledge  (HILK) methodology, they ``are able to represent the extracted knowledge in highly interpretable fuzzy rule-based systems'' \cite{conde2013interpretable}. HILK is a fuzzy modeling methodology intended to produce classifiers easily comprehensible by humans. As in the previously mentioned fuzzy logic approaches, model interpretability is not measured but assumed to be a natural result of the methodology application, where a set of conditions must be met at design time.

\textbf{Machine learning} has recently become the most active field of research on interpretability. More interpretable models are being pursued not only for more complex techniques such as deep neural networks but also for more traditional models, like the interpretable \textit{decision lists} proposed by Letham et al. \citeyear{letham2013interpretable}. A decision list is a series of \textit{if-then-else} statements, where the \textit{if} statements define a partition of a set of features, the \textit{then} statements define a predicted outcome, and the \textit{else} statements define either a new rule to be applied (if followed by another \textit{if} statement) or a default outcome to be assumed as the prediction, in case none of the previous rules were assessed as true. The proposed approach, called Bayesian List Machine (BLM), ``produces a posterior distribution over permutations of \textit{if...then...} rules, starting from a large set of possible pre-mined rules'' \cite{letham2013interpretable}. The authors follow the assumption that the model is intrinsically interpretable because, given their format, the rules naturally provide a reason for the prediction they lead to. However, no quantitative or qualitative evaluation is presented regarding the model's interpretability assessment.

\begin{figure*}[t]
	\centering
	\includegraphics[width=\textwidth]{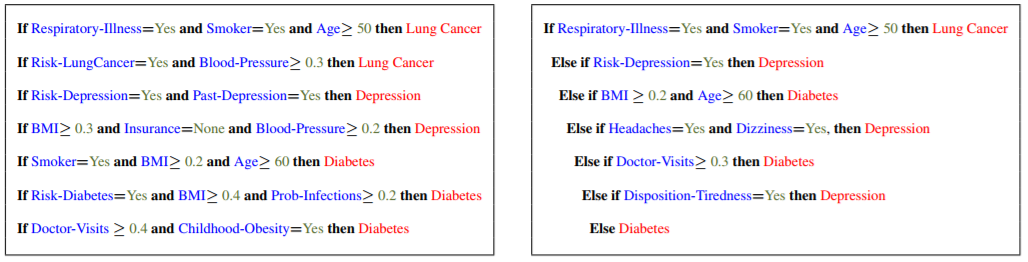}
	\caption{A decision set (left) and a decision list (right) learned from a diagnosis dataset, as provided by Lakkaraju et al. \protect\citeyear{lakkaraju2016interpretable}. Decision set rules can be interpreted independently, while in decision lists every rule depends on all the rules above it.}

	\label{fig:set_list}
\end{figure*}

Lakkaraju et al. \citeyear{lakkaraju2016interpretable} introduced a predictive model based on interpretable \textit{decision sets} (DSs): sets of independent \textit{if-then} rules aimed at being human-interpretable while still showing high accuracy. They define decision sets as ``a model class that can both accurately predict class labels and interpretably describe its decision boundaries'' \cite{lakkaraju2016interpretable}, and claim they are more interpretable than decision lists because the if-then rules, organized in a non-hierarchical structure, apply independently and can be considered in any order. On the other hand, the rules in decision lists are in the if-then-else format, and each rule depends on all the rules above it, being necessary to interpret the whole hierarchical structure to understand why a given rule is applied. This difference is exemplified in Figure \ref{fig:set_list}. Decision sets are concerned only with model transparency, and the authors measure the model interpretability both quantitatively (see the Evaluation Methods and Initiatives Section) and through a qualitative evaluation, which measured to what extent human subjects could interpret the rules, that is, determine which decision each rule was leading to \cite{lakkaraju2016interpretable}.

Aiming at making statistical modeling accessible to non-experts, Lloyd et al. \citeyear{lloyd2014automatic} introduces the Automatic Statistician, a system that analyses data sets and automatically discovers the statistical model describing the data, generating a report with figures and natural language text. The proposed approach, called Automatic Bayesian Covariance Discovery (ABCD), focus on Nonparametric Regression Models and uses a greed search procedure to explore the space of regression models, finding, through Bayesian inference, the main components describing the data, such as ``periodic function'', ``linear function'', ``constant'', ``smooth function'', among others. The discovered model's components are then translated into English phrases, resulting in a report with text, figures, and tables, detailing what has been inferred about the data, besides model checking and criticism \cite{ghahramani2015probabilistic}. The final report allows users to interpret the system decisions without having to go through its internal operations, still, qualitative interpretability evaluation was not presented.

As well as the ABCD approach, the technique proposed by Ribeiro et al. \citeyear{ribeiro2016should} can be seen as an \textit{interpretability layer} for machine learning models, in the sense that they do not try to adjust the algorithms in order to make them more interpretable, but rather analyze their outputs to build an explanation around it. The Local Interpretable Model-agnostic Explanations (LIME) approach can do so for any classifier by learning an interpretable model that approximates a prediction locally \cite{ribeiro2016should}. The explanations consist of a set of artifacts that were relevant for the model's prediction, be it textual (for example, which words were decisive in a text classification task) or visual (for instance, which elements present in a picture influenced an image classification task). These artifacts can assume any representation format and can be seen as an abstraction for the features used by the model, since these features can be too complex to be presented to the user (like word embeddings, for example). The explanation for a prediction is built by sampling instances in its vicinity and selecting the features that were relevant for those instances classification. This method allows the explanation not only of a single prediction, but also of the whole model, by selecting a set of representative instances, a method called \textit{submodular pick}, and presenting the explanations for them. This allows the users to have an overall insight into how decisions are being made and decide whether they can trust them or not, while the model itself can still be treated as a black box. Quantitative and qualitative evaluations (see the Evaluation Methods and Initiatives Section) show good results for text and image classification tasks. However, what a suitable interpretable representation for an explanation is and how complex is to derive such representation will depend on the model and on the task being addressed.

The Gray Box Decision Characterization (GBDC) approach \cite{brinton17framework} also uses some knowledge about the model's internal procedures to generate post-hoc explanations, without modifying the model itself. That means it could also be used as an interpretability layer for any classifier or regression model. It focuses on characterizing, i.e., explaining a single prediction at a time by performing a sensitivity analysis around its input data vector. The GBDC searches for changes in the basis vector contained in the space region around the input data that lead to changes in the model's output. The explanation for a prediction is then constructed by selecting the features that yield the most significant changes in this specific output. Explanations are provided in natural language but, as the author points out, interpretability evaluation including human subjects is yet to be done.

Datta et al. \citeyear{datta2017algorithmic} go one step further and, besides pointing out which features most likely led to a decision, measure the influence of each of these features on the prediction. By defining a family of Quantitative Input Influence (QII) measures, transparency queries can be posed to the model so decisions about individuals and groups can be explained. \textit{Joint influence} of a group of features and \textit{marginal influence} of individual inputs can also be measured when single inputs do not have high influence. Like the GBDC approach, QII forces changes on the inputs to check whether they lead to changes in the output for identifying influential features, but also attributes weights, placing the features in a rating scale that shows clearly how influential each feature was for the decision. The QII measures can also be applied to any classifier and are used to produce \textit{transparency reports}, which, for a given prediction, shows the QII measures for every input feature in graph format. The transparency reports allow for a clear interpretation of how the decision was reached and make it easier spotting spurious correlations, but still need to be qualitatively evaluated by human users.

Rather than performing post-processing to interpret some model's results, the Mind the Gap Model (MGM) opts for embedding interpretability criteria directly into the model \cite{kim2015mind}. The MGM is a generative approach for feature extraction and selection which aims at identifying not only what features characterize a cluster, but also what features distinguish between clusters. A logic-based feature extraction consolidates dimensions into groups, followed by the identification of important groups based on parameter values, which selects groups having gaps in their parameter values across clusters. The feature groups consist of logical formulas governed by either the \textit{or} or the \textit{and} logical operators, and each group is associated to a cluster through a probability value, which indicates how likely the features in the group are to appear in the cluster. This \textit{feature group} vs. \textit{cluster} matrix is presented to the user as an explanation for the final data clustering. Qualitative evaluation including the participation of human subjects to verify the model interpretability has shown that domain experts could easily understand and quickly write an executive summary of this matrix, as well as finding the differences between clusters.

\begin{figure*}[t]
	\centering
	\includegraphics[width=\textwidth]{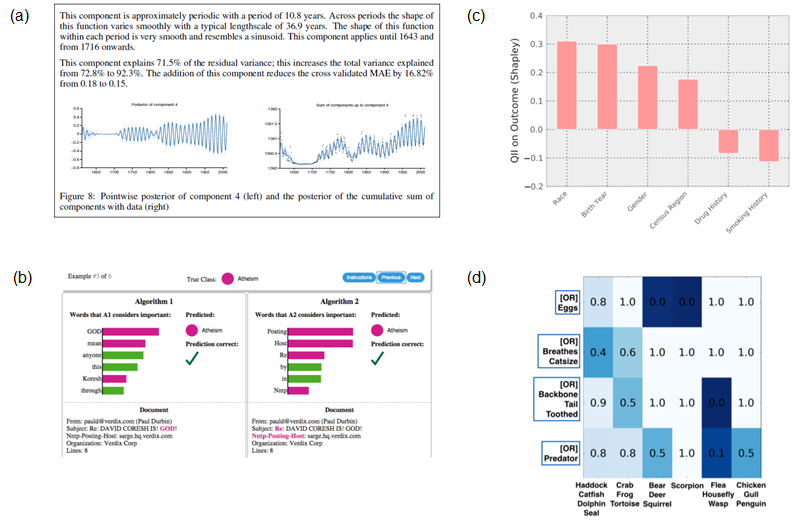}
	\caption{Examples of post-hoc explanations: (a) ABCD report \protect\cite{lloyd2014automatic}; (b) LIME prediction explanation \protect\cite{ribeiro2016should}; (c) QII transparency report \protect\cite{datta2017algorithmic}; and (d) MGM feature matrix \protect\cite{kim2015mind}.}

	\label{fig:reports}
\end{figure*}

Figure \ref{fig:reports} shows some examples of post-hoc explanations produced by the ABCD, LIME, QII, and MGM approaches. ABCD reports rely on natural language descriptions and graphs to describe the discovered data models. LIME shows the features that influenced a prediction in a classification task and QII does the same but placing the features on a numerical scale. MGM presents groups of features along with their likelihoods of belonging in each cluster.

\subsection{Hybrid Models}
Interpretable hybrid models are a mix of data and algorithmic models. Besides having a data component, they also include an algorithmic procedure which makes use of this data for a predefined task, meaning that the data component is designed and created specifically to suit the algorithm goals.

\textbf{Topic models} can be seen as hybrid models, as they encompass both a data model -- the set of topics, which are collections of words similar to the DSM's dimensions -- and an algorithmic procedure, used to classify the documents according to the previously discovered topics. Topic modeling algorithms can be applied to large and unstructured collections of documents, discovering their main themes and categorizing their documents based on the discovered themes \cite{blei2012probabilistic}. Formally, probabilistic topic models are resources composed by a set of latent topics for performing the unsupervised analysis of large document corpora, and assume that each document in the collection can be described by a combination of such topics \cite{chang2009reading}.

Although it is usually assumed that the resulting semantic space is always meaningful, its interpretability can't be measured by the commonly used predictive evaluation metrics, that is, quantitative metrics that capture the model's ability to predict the topics for unseen documents. Chang et al. \citeyear{chang2009reading} show that by evaluating topic models generated by three different techniques: Latent Dirichlet Allocation (LDA) \cite{blei2003latent}, Probabilistic Latent Semantic Indexing (pLSI) \cite{hofmann1999probabilistic} and Correlated Topic Model (CTM) \cite{blei2005correlated}, through the word intrusion and topic intrusion tasks (see the Evaluation Methods and Initiatives Section). They show that LDA presents the best results for the word intrusion task, and is comparable to pLSI for the topic intrusion one, while CTM showed to be the less interpretable model, despite presenting the highest predictive likelihood among the three. This leads to the conclusion that the highest probability does not entail the best interpretability. As the trend was, in fact, the opposite, this could suggest that as topics become more fine-grained in models with a larger number of topics, they are less useful for humans \cite{chang2009reading}.

For learning semantically consistent topics right from the beginning, topic models are starting to be designed with the explicit goal of being interpretable. Among the models that claim to favor interpretability is the Topical N-Grams (TNG) \cite{wang2007topical}, a topic model that takes into account the order of words in text to discover topical phrases, in contrast to models such as LDA that generates topics under the bag-of-words assumption, that is, assuming that words are generated independently from each other. The authors argue that whether or not a phrase is a collocation may depend on the topic context and that the TNG is capable of making that distinction. TNG proves to be more interpretable than LDA especially when dealing with generic words, such as ``state'' or ``action'', which, alone, may seem misplaced in a topic for being too vague (when interpreted by a human), but are far more meaningful when forming n-grams such as ``belief state'' or ``action selection'' in a reinforcement learning-themed topic, for example.

Ramage et al. \citeyear{ramage2011partially} also claim that their models, Partially Labeled Dirichlet Allocation (PLDA) and Partially Labeled Dirichlet Process (PLDP), are more interpretable than unsupervised approaches. PLDA and PLDP combine unsupervised machine learning-based discovery of topics with content annotated with human-provided labels. PLDA ``is a generative model for a collection of labeled documents, extending the generative story of LDA to incorporate labels, and of Labeled LDA to incorporate per-label latent topics'' and PLDP ``replaces PLDA's per-label topic mixture [...] with a Dirichlet process mixture model'' \cite{ramage2011partially}. Both models learn the topic structure within the scope of the observed labels, which impose a kind of semantic constraint on the resulting model. However, although they argue that this constraint improves correlation with similarity judgments, the case studies presented involves no human participation. Tying discovered topics to human interpretable labels can indeed produce a more interpretable topic structure, but how humans actually evaluate both the topic's consistency and the adequacy of the associations between topics and documents provided by these two models remains to be measured.

The work introduced by Song et al. \citeyear{song2011short} is a kind of topic modeling focused on short texts, such as Twitter messages. Since such short texts don't provide enough contextual information, the topics can't be derived solely from their content. Instead, the words in the text are mapped to entities, which can be either concepts (e.g. ``country''), instances of concepts (e.g. ``China'', ``India'') or attributes of concepts/instances (e.g. ``language'', ``population''), in a probabilistic knowledge base called \textit{Probase}. Those entities, along with the relationships between them, will then provide the context for the topic discovery, which the authors call \textit{conceptualization}, i.e., the definition of a set of concepts that best describe the text's content. The conceptualization is carried out through Bayesian inference (BI), benefiting from the large size of the database. The quantitative evaluation, which uses clustering-related measures, such as purity, adjusted random index (ARI) and normalized mutual information (NMI), shows the approach yields good accuracy but, although the resulting topics presented as examples are in fact easily interpretable, no evaluation regarding the model interpretability itself was performed.

Table \ref{tab:models} summarizes the main characteristics of the analyzed data, algorithmic and hybrid models.

\begin{table*}[t]
	\centering
	\caption{Analyzed interpretable AI models. DSM = Distributional Semantics Model; FL = Fuzzy Logic Model; ML = Machine Learning Model; TM = Topic Model; PM = Probabilistic Model. The Presentation column indicates how the model transparency and/or post-hoc explanation is presented to the user.}
	\label{tab:models}
	\begin{tabularx}{\textwidth}{lcccccc}
		\multicolumn{1}{c}{\textbf{Approach}} & \textbf{Model} & \textbf{Type} &  \textbf{Transparency} & \textbf{Explanation} & \textbf{Presentation} \\ \hline
		Strudel \cite{baroni2010strudel}         
		& DSM            & data                &  yes                         & no                            & textual               \\
		NNSE \cite{murphy2012learning}            
		& DSM            & data                & yes                         & no                            & textual               \\
		CNNSE \cite{fyshe2015compositional}            
		& DSM            & data                &  yes                         & no                            & textual               \\
		JNNSE \cite{fyshe2014interpretable}            
		& DSM            & data                &  yes                         & no                            & textual               \\
		EPCA \cite{brinton17framework}                  
		& DSM            & data                &  yes                         & no                            & textual               \\
		MP \cite{balazs2013stochastic}              
		& FL             & algorithm           &  yes                         & no                            & textual               \\
		SFP \cite{mencar2013design}            
		& FL             & algorithm           &  yes                         & no                            & textual               \\
		FAR-Fingrams \cite{pancho2013interpretability}    
		& FL             & algorithm         &  yes                         & no                            & visual                \\
		HC \cite{riid2013determination}                
		& FL             & algorithm          & yes                         & no                            & textual/visual        \\
		HILK \cite{conde2013interpretable}    
		& FL             & algorithm           & yes                         & no                            & textual               \\
		DS \cite{lakkaraju2016interpretable}           
		& ML             & algorithm            & yes                         & no                            & textual               \\
		BLM \cite{letham2013interpretable}             
		& ML             & algorithm           & yes                         & no                            & textual               \\
		ABCD \cite{lloyd2014automatic}             
		& ML             & algorithm            & no                          & yes                           & textual/visual        \\
		LIME \cite{ribeiro2016should}           
		& ML             & algorithm           & no                          & yes                           & textual/visual        \\
		GBDC \cite{brinton17framework}                  
		& ML             & algorithm           & no                          & yes                           & textual               \\
		QII \cite{datta2017algorithmic}              
		& ML             & algorithm            & no                          & yes                           & visual                \\
		MGM \cite{kim2015mind}                
		& ML             & algorithm            & yes                           & 
		yes	                          & textual               \\
		TNG \cite{wang2007topical}               
		& TM             & hybrid               & yes                         & no                            & textual               \\
		PLDA \cite{ramage2011partially}            
		& TM             & hybrid              & yes                         & no                            & textual               \\
		PLDP \cite{ramage2011partially}            
		& TM             & hybrid               & yes                         & no                            & textual               \\
		BI \cite{song2011short}                
		& PM             & hybrid               & yes                         & no                            & textual              
	\end{tabularx}
\end{table*}

\subsection{Evaluation Methods and Initiatives}
Despite the crescent effort to build more interpretable models, measuring their level of interpretability is still a challenge, regardless of the type of the model. Some evaluation methods have been proposed, but always in the context of a specific model, whose characteristics determine what will be assessed and which measures are relevant for the evaluation.

An initiative towards the qualitative evaluation of topic models is the set of tasks proposed by Chang et al. \citeyear{chang2009reading}, intended to measure the model's interpretability. In the \textit{word intrusion} task, a human evaluator needs to find the word that does not belong in the topic, i.e., the \textit{intruder}, assessing the model's data component. The algorithmic component is evaluated through the \textit{topic intrusion} task, which checks whether the topics assigned to a document by a topic model match the human judgments of the document's content, by presenting a document and a set of topics related to it to a human evaluator and asking them to identify the intruder topic. The underlying rationale is the same for both tasks: if the topic is semantically consistent, that is, if the words that compose it refer to the same semantic category, even if in a broad sense, then humans can easily interpret its meaning and point out what does not belong to it, be it a single word or a whole document. The word intrusion task is also commonly used to assess the semantic coherence of DSM dimensions \cite{murphy2012learning,fyshe2015compositional}.

In the fuzzy logic context, evaluating model interpretability is not yet a widespread practice. As seen in the Interpretability-driven AI Models Section, most fuzzy models claim to be interpretable because they meet a set of constraints at design time, but no quantitative or qualitative evaluation is in fact carried out so the model interpretability can be assessed from a human point of view. Despite that, a large number of objective and subjective indexes for assessing the interpretability of fuzzy systems have been proposed, covering different granularity levels inside the model \cite{alonso2015interpretability}. These indexes are intended to analyze both the structural-based interpretability and semantic-based interpretability of a model, and takes into account a set of constraints and criteria regarding each of the model's abstraction layers, from the lowest to the highest: fuzzy sets, fuzzy partitions, fuzzy rules and fuzzy rule bases. Structural constraints and criteria refer to the internal organization of the elements that compose a fuzzy model, and determine its \textit{readability} level, while semantic constraints and criteria refer to the way the fuzzy system behaves, that is, how it reaches its results, dictating the model's \textit{comprehensibility} level. A detailed list of constraints and criteria, as well as a description of the most outstanding interpretability indexes for evaluating fuzzy models is provided by \cite{alonso2015interpretability}.

Mencar et al. \citeyear{mencar2011interpretability} go in a different direction and propose an interpretability evaluation method using the notion of \textit{cointension}, defined as ``a measure of proximity of the input/output relations of the object of modeling and the model'' \cite{mencar2011interpretability}. Exploiting the \textit{logical view}, a set of properties expected to contain the (approximated) implicit semantics and defined as ``the propositional structure of the rules in the knowledge base, responding to the laws of formal logics both for the fuzzy rule-based inference and the user thinking'' \cite{mencar2011interpretability}, the rule base is transformed into a different one logically equivalent to it. The model interpretability is then measured by the extent to which the retained logical equivalence of the rule bases corresponds to the semantic equivalence. In practice, this is a quantitative assessment of interpretability: the model is interpretable if the logical view of its rules is correct, what is measured by the comparison between the accuracies of the two rule bases (the lower the difference, the higher the logical view correctness and, consequently, the interpretability).

Regarding machine learning models, Doshi-Velez and Kim \citeyear{doshi2017roadmap} go in a more theoretical direction and propose, rather than a method, a taxonomy for the evaluation of model interpretability. They observe that current evaluation techniques either assesses interpretability in the context of an application or via a quantifiable proxy. In the first case, a system is considered interpretable if it is useful in a practical application (or a simplified version of it); in the second scenario, a model is considered intrinsically interpretable and is just subjected to optimization algorithms. They argue that, although these approaches may seem, at first, reasonable, they lack rigor and, in order to compare methods in an effective way, evaluation criteria must be formalized and based on evidence. Hence, in the proposed taxonomy the evaluation method, which should be based on sets of task- and method-related latent dimensions, may vary from model to model and must take into account the focus of the contribution, which can range from assessing the reliability of a real-world application from the human point of view to better optimizing a given model with regard to interpretability requirements.

In a more practical fashion, Lakkaraju et al. \citeyear{lakkaraju2016interpretable} propose, in addition to asking human subjects to relate a given decision to the rules that generated it, a quantitative evaluation for assessing decision sets' interpretability. They define four metrics: \textit{size} (the number of rules in the set), \textit{length} (the size of each rule), \textit{cover} (the number of points in the data set covered by each rule) and \textit{overlap} (the number of features covered by more than one rule). Under this evaluation framework, the lowest the size, length and overlap of a decision set, the highest its interpretability (cover is used as an intermediary metric to compute overlap).

The evaluation methods introduced by Ribeiro et al. \citeyear{ribeiro2016should} also target machine learning models, but focus on assessing the interpretability of the post-hoc explanations generated for their predictions. The set of evaluation tasks includes both quantitative and qualitative assessments. The quantitative tasks aims at determining if the explanations are faithful to the model, if a single prediction is trustworthy, and if the model as whole is trustworthy. As the explanations are basically composed of a set of features, this is done by creating gold standard sets of features and computing the model's recall for each explanation it generates (for measuring faithfulness); by marking a subset of features as ``untrustworthy'' and measuring the rate of such features in an explanation (for computing the prediction's trustworthiness); and by adding artificial ``noisy'' features to the model to evaluate how many of its predictions can be trusted, i.e., how many predictions do not include the untrustworthy noisy features in their explanations (for measuring the model's overall trustworthiness). 

The qualitative assessment is carried out by showing human subjects the explanations generated by two classifiers and measuring to what extent these explanations help them to make decisions. First, users need to select the best classifier based on the explanations they provide for a text classification task, where one classifier uses untrustworthy features (words) and the other one (considered the best one) does not. Second, also in the context of a text classification task, (non-expert) users need to improve a classifier, by analyzing the explanations and removing the features (words) they consider untrustworthy for subsequent model training. Using an image classification task as context, the third evaluation task forces a wrong correlation by selecting all images from a class having a given feature that does not generalize in the real world (in this case, the classes used were ``wolf'' and ``husky'', and all the wolf pictures contained snow, which would end up being used by the classifier for generalization). The objective is to evaluate if observing the explanation for a prediction, which in this case is a super-pixel of the image, users can have insights on which features are being used by the model, identify if it is making spurious correlations and decide whether it can be trusted or not. The rationale behind the three tasks is the same: if human users make the expected decisions, it can be concluded that the explanations are in fact allowing the correct model interpretation.

A few evaluation challenges have also been realized to stimulate the addition of interpretability features in semantic applications. The third PASCAL Recognizing Textual Entailment Challenge (RTE-3) included an optional task requiring the participant system to justify their answer, that is, besides deciding whether a piece of text (the entailed \textit{hypothesis}) could be entailed from another one (the entailing \textit{text}) or not, they should also provide a post-hoc explanation justifying this decision \cite{voorhees2008contradictions}. The explanations, which could be a collection of strings of any size, should be given in terms suitable for an end user (i.e., not a system developer), and were evaluated by human judges who assessed their understandability and correctness. Some common concerns and criticisms in the judges' evaluation summaries include verbatim repetition of the text and hypothesis in the justification, use of generic phrases such as ``there is a relation'' and ``there is a match'', presentation of system internals such as numerical similarity scores, and use of mathematical notation and linguistic jargon such as ``polarity'' and ``hyponym''. These observations point out that conciseness and specificity are important features from the user point of view, and must have priority when explaining the system behavior.

Also in the Natural Language Processing field, an interpretable semantic text similarity (STS) task was proposed at SemEval \cite{agirre2015semeval}. Participants were asked, in addition of rating the degree of semantic equivalence between two text snippets, to include an explanatory layer responsible for aligning the chunks in the sentence pair and annotating the kind of relation and the similarity score of the chunk pair. Rather than providing a natural language (post-hoc) explanation, systems should only justify the overall similarity score by pointing which parts in both pieces of text contributed to this score. Two scenarios were proposed: in the first one, participants were given gold standards chunks and should only make the correct alignments and provide them with appropriate labels and scores. In the second scenario they were given raw text as input, and should also segment the input. The relevant relations defined for the task include \textit{EQUI} (semantically equivalent), \textit{OPPO} (oppositional meaning), \textit{SPE1/SPE2} (chunk in sentence 1 is more specific than the one in sentence 2 and vice-versa), \textit{SIM} (similar meaning, other than EQUI, OPPO, and SPE1/SPE2), and \textit{REL} (related meaning, other than EQUI, OPPO, SPE1/SPE2, and SIM). Other relations refer to non-aligned chunks (\textit{NOALI}), or context alignments (\textit{ALIC}), that is, chunks that should be aligned to a chunk that was already aligned previously but can't due to a 1:1 alignment restriction. Since gold standards for chunk alignments, relations and similarity scores were available for all the text pairs, the evaluation was purely quantitative, measuring the F1 score obtained by each participant system. From the interpretability point of view, the output produced by the systems can't be assessed individually, since they followed the task requirements, providing only chunk alignments, with a relation from a predefined set and a similarity score associated with each alignment. Regarding the predefined relationships, SIM and REL relations seem particularly vague: the difference between them is not clear and both can refer to any semantic relation other than equivalence, opposition, and hypernymy, without ever making explicit what this relation is. The task may doubtless have served as a first exercise towards extracting further information for allowing a system interpretation, but turning this information into useful explanation for end users would require subsequent data refinement and formatting. 

Table \ref{tab:eval} summarizes the main characteristics of the aforementioned interpretability evaluation tasks.

\begin{table*}[t]
	\centering
	\caption{Interpretability evaluation tasks. On the Type column \textit{qual} stands for qualitative and \textit{quant} for quantitative evaluation.}
	\label{tab:eval}
	\begin{tabular}{lccccc}
		\multicolumn{1}{c}{\thead{Task}}                   & \thead{Type}            & \thead{Target Model} & \thead{Element \\ Evaluated} & \thead{Dimension  \\ Evaluated}   & \thead{Human \\ Evaluation} \\ \hline
		\makecell[l]{Word intrusion \\ \cite{chang2009reading}}                & qual & \makecell{TM, DSM}               & model                      & consistency                    & yes                       \\
		\makecell[l]{Topic intrusion \\ \cite{chang2009reading}}                & qual & TM                    & model                      & consistency                    & yes                       \\
		\makecell[l]{Logical view equivalence \\ \cite{mencar2011interpretability}}      & quant             & FL                    & model                      & cointension                    & no                        \\
		\makecell[l]{Rule interpretation \\ \cite{lakkaraju2016interpretable}}        & qual              & ML (DS)               & model                      & understandability              & yes                       \\
		\makecell[l]{Decision set measuring \\ \cite{lakkaraju2016interpretable}}     & quant             & ML (DS)               & model                      & \makecell{size, length, \\ cover, overlap}   & no                        \\
		\makecell[l]{Feature recall assessment \\ \cite{ribeiro2016should}}    & quant             & ML                    & explanation                & faithfulness                   & no                        \\
		\makecell[l]{Feature precision assessment \\ \cite{ribeiro2016should}} & quant             & ML                    & explanation                & trustworthiness                & no                        \\
		\makecell[l]{Feature noise insertion \\ \cite{ribeiro2016should}}      & quant             & ML                    & model                      & trustworthiness                & no                        \\
		\makecell[l]{Classifier selection \\ \cite{ribeiro2016should}}         & qual              & ML                    & explanation                & understandability              & yes                       \\
		\makecell[l]{Classifier improvement \\ \cite{ribeiro2016should}}       & qual              & ML                    & explanation                & understandability              & yes                       \\
		\makecell[l]{Feature identification \\ \cite{ribeiro2016should}}       & qual              & ML                    & explanation                & understandability              & yes                       \\
		\makecell[l]{RTE justification \\ \cite{voorhees2008contradictions}}                  & qual              & n/a                   & explanation                & \makecell{understandability,  \\ correctness} & yes                       \\
		\makecell[l]{STS explanation \\ \cite{agirre2015semeval}}               & quant             & n/a                   & explanation                & accuracy                       & no                       
	\end{tabular}
\end{table*}

\section{A Human-Centered Take on Interpretability}
Given the heterogeneity of the models tackling interpretability, the product offered as an interpretation and the point in the system's workflow at which this is accomplished also widely vary. Both aspects can affect the way the final user benefits from an interpretable model. It may not be enough for a model being interpretable; model interpretability should be incorporated seamlessly into the user's routine while using the system as a support tool.

\begin{figure}[ht]
	\centering
	\includegraphics[]{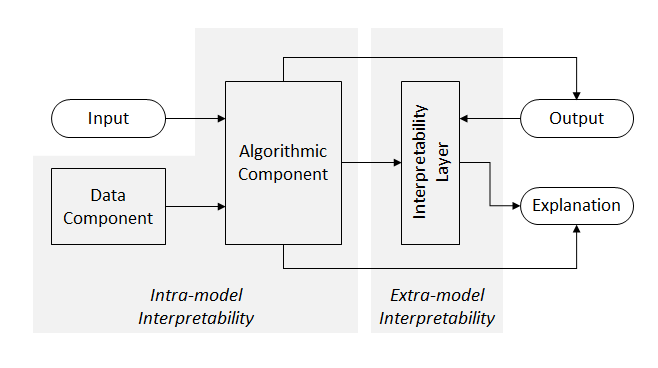}
	\caption{A generic AI system architecture and the points where interpretability features can be inserted.}
	\label{fig:architecture}
\end{figure}

To visualize the differences between approaches, consider the generic system architecture depicted in Figure \ref{fig:architecture}, which sums up the characteristics of the various models described in the Interpretability-driven AI Models Section. In this architecture, an input is sent to an algorithmic component, which will perform a sequence of computations over it and produce an output. Optionally, the algorithm can also employ the features from a data component, which can be an external model developed independently (such as a DSM) or an internal component tailored to the algorithm needs (the set of topics in a topic model, for instance). Besides the output itself, the algorithm may also provide an explanation for it. Alternatively, this explanation can be produced by an interpretability layer, a component that will take the output and induce the model behavior that generated it.

Interpretability features at the data and algorithmic component levels can be considered \textit{intra-model}, that is, the interpretability is embedded in the model and influences the way it works in order to keep its behavior understandable from the user point of view. On the other hand, an interpretability layer can be seen as an \textit{extra-model} component, since it usually is model-independent and tries to explain the algorithm behavior without modifying it or interfering in the way it works.

Intra-model interpretability naturally favors transparency, but can also work on the generation of post-hoc explanation if the algorithmic component is designed for providing complementary information to the predicted output. Extra-model interpretability is an effective tool for providing comprehensible post-hoc explanations, but at the expense of transparency, since it usually interprets the model locally (around the output being explained); interpreting the whole model is a challenging task, so, globally, it may remain a black box.

\begin{figure}[ht]
	\centering
	\includegraphics[]{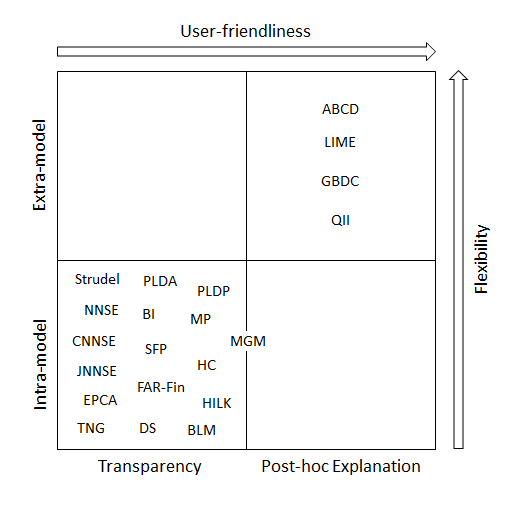}
	\caption{The four quadrants of AI models' interpretability.}
	\label{fig:quadrants}
\end{figure}

In practice, most models that implement intra-model interpretability will be contented with transparency, doing without any kind of post-hoc explanation. Figure \ref{fig:quadrants} shows this trend, situating the models analyzed in the Interpretability-driven AI Models Section according to the points where they introduce interpretability features and the kind of interpretation they offer. As can be seen, the majority of the models offers only transparency through intra-model interpretability, while a few provide only post-hoc explanations making use of extra-model interpretability features. The sole exception is MGM \cite{kim2015mind}, which provides post-hoc explanations while still claiming to be a transparent, interpretable model.

Taking into account the aforementioned human-centric nature of interpretability, this classification allows us to identify two important dimensions: the ease of use of interpretability features and the impact of their introduction from the point of view of the final user. First, it may be easy to understand the internal operations of a transparent model, but not all users will be willing to do so. A machine learning engineer seeking to tune a neural network will be happy to track the flow of computations of the model; a physician looking for the evidence for a diagnosis or a credit analyst who needs to justify a denied loan probably won't. Post-hoc explanations, either in textual or visual formats, tell how the model is working in a user-friendlier way and are better suited for non-developer users.

Second, new, interpretable versions of already existing models are important additions to the artificial intelligence body of knowledge, but, given the current widespread use of such technologies, ditching a deployed, fully functional system altogether for a new explainable one is not always an option. Despite the efforts in the direction of preserving the model performance while making it interpretable, the tradeoff between interpretability and accuracy is still an issue \cite{balazs2013stochastic}. Interpretability layers can offer higher flexibility while evolving a system, since they can be attached to any model, and without affecting the way it is already working, ensuring that results won't change due to newly introduced interpretability functionalities. For large scale and/or critical systems, adding such layers represent a low-risk incremental upgrade, potentially meaning also a low impact on the final user routine.

Increasing model transparency is an important step towards widespread Explainable AI, but it should be only the first one if the user needs are put in the first place. A transparent model already offers a rich material for providing the user with explanations, requiring only a translation step to format the sequence of interpretable operations that led to an output as natural language justifications, image highlighting, graphs, diagrams or any other form of communication suited to the task at hand, as long as it supports the user's work instead of adding further workload. Any model can benefit from an interpretability layer; even though fuzzy sets or decision lists are more comprehensible than deep neural networks, they still carry logic formalisms, which the user can be spared of. Given the wide variety of available artificial intelligence approaches and tasks to be addressed, generating post-hoc explanations, either at the intra- or extra-model level, is still an open challenge, and a path worth being explored even by the already interpretable (transparent) models.

\section{Conclusion}
Artificial intelligence algorithms can now process huge amounts of data and draw conclusions from it in a way not attainable for humans, making a wide variety of tasks quicker and easier than ever. However, either supporting users' decisions or automating those decisions for them, any AI system will be attached to a human who will be responsible for its decisions and the consequences these decisions may entail. Humans need to explain and justify their acts, and so do the algorithms that will take a substantial part in humans' decision-making.

After successful efforts for making AI models highly accurate, researchers are now being faced with the challenge of making them also interpretable, untangling their complexity to make the rationale behind any prediction clear and intelligible. By understanding the model behavior and being able to explain its decisions, users can not only justify the decisions they make based on it, but also identify when it is making spurious correlations or reflecting social biases contained in data to avoid unfair decisions.

In this work we sought to look at semantic interpretability as a cross-field subject, going beyond the machine learning perspective and bringing to light the efforts of other areas, such as distributional semantics and fuzzy logic, towards increasing model interpretability. By examining how distinct disciplines define and offer interpretability, we outlined the shapes it can assume and, based on them, analyzed several different types of models and interpretability evaluation methods. We further categorized the models according to how they integrate interpretability features into their architectures and assessed how this, along with the type of interpretation offered, impacts the final user routine.

Despite many advancements have been reported recently, given the wide variety of models implementing AI functions, there is still much work to be done. Providing interpretability for a broad range of diverse models, keeping explanations simple, useful and user-friendly while integrating them in the user's environment without affecting current results and overall procedures is an open challenge that will require researchers to switch from accuracy-only driven approaches to human-centered model analysis, design, and implementation choices.

\bibliography{interpretability_survey}
\bibliographystyle{apacite}

\end{document}